# Enhanced Cooperative Perception Through Asynchronous Vehicle to Infrastructure Framework with Delay Mitigation for Connected and Automated Vehicles

Nithish Kumar Saravanan [1], Varun Jammula [1], Yezhou Yang [1], Jeffrey Wishart [2] and Junfeng Zhao [1]

[1] Arizona State University
[2] Science Foundation Arizona
{nsarava3, vjammula, yz.yang}@asu.edu, jeffw@AZcommerce.com, jzhao176@asu.edu

*Abstract*— Perception is a key component of Automated vehicles (AVs). However, sensors mounted to the AVs often encounter blind spots due to obstructions from other vehicles, infrastructure, or objects in the surrounding area. While recent advancements in planning and control algorithms help AVs react to sudden object appearances from blind spots at low speeds and less complex scenarios, challenges remain at high speeds and complex intersections. Vehicle to Infrastructure (V2I) technology promises to enhance scene representation for AVs in complex intersections, providing sufficient time and distance to react to adversary vehicles violating traffic rules. Most existing methods for infrastructure-based vehicle detection and tracking rely on LIDAR, RADAR or sensor fusion methods, such as LIDAR-Camera and RADAR-Camera. Although LIDAR and RADAR provide accurate spatial information, the sparsity of point cloud data limits its ability to capture detailed object contours of objects far away, resulting in inaccurate 3D object detection results. Furthermore, the absence of LIDAR or RADAR at every intersection increases the cost of implementing V2I technology. To address these challenges, this paper proposes a V2I framework that utilizes monocular traffic cameras at road intersections to detect 3D objects. The results from the roadside unit (RSU) are then combined with the on-board system using an asynchronous late fusion method to enhance scene representation. Additionally, the proposed framework provides a time delay compensation module to compensate for the processing and transmission delay from the RSU. Lastly, the V2I framework is tested by simulating and validating a scenario similar to the one described in an industry report by Waymo. The results show that the proposed method improves the scene representation and the AV's perception range, giving enough time and space to react to adversary vehicles.

GitHub repository: https://github.com/BELIV-ASU/Enhanced-Cooperative-Perception-Through-Asynchronous-V2I-Framework-with-Delay-Mitigation-for-CAVs

## I. Introduction

Automated vehicle (AV) technology has evolved significantly over the past decade. The advancement of high-resolution LIDARs, cameras, and computing units, along with high-speed and accurate perception algorithms, has significantly contributed to this tremendous change. Despite the high-level accuracy of the perception systems available today, blind spots due to occlusion cannot be overcome by the sensors mounted on AVs due to their limited field of view and location on the AV. Occlusions can be created by nearby vehicles, infrastructure, or other structural elements. Vehicle-to-Infrastructure (V2I) communication is a promising technology to overcome occlusion challenges in complex road networks.

Numerous roadside unit (RSU)-based LIDAR perception systems have shown capability in delivering accurate spatial information [1, 2, 3]. Algorithms for 3D object detection using traditional machine learning techniques with LIDAR data typically incorporate background filtering to eliminate irrelevant points, thereby preserving only those points associated with objects [4, 5, 6]. The filtered foreground points are clustered utilizing techniques such as Euclidean clustering and DBSCAN clustering [7, 8]. Clustered point clouds are classified employing traditional classifiers, including Support Vector Machines (SVM), decision trees, and Artificial Neural Networks (ANNs). Traditional 3D LIDAR object detection methods utilizing machine learning often encounter challenges in selecting appropriate thresholds for filtering and clustering. Additionally, their performance may be compromised in dynamic environments due to the algorithm's reliance on fixed features such as length, height, and length-height ratio used in background filtering and clustering. Algorithms for 3D object detection using deep learning techniques surpass the constraints of traditional machine learning algorithms by facilitating automatic feature extraction via data-driven learning [9, 10, 11]. The LIDAR point cloud sparsity significantly limits the geometric information of faraway objects, lowering detection accuracy and range compared to camera-based perception systems. LIDAR-camera fusion systems improve classification accuracy; however, they do not considerably extend the detection range [12]. The higher cost of LIDAR systems poses challenges for the implementation of RSUs in V2I communication.

Conversely, few RADAR-based methods use adapted versions of LIDAR algorithms for 3D object detection [13, 14]. Nonetheless, the RADAR-based approach suffers from limitations such as detecting stationary objects as clutter and losing geometric information of faraway objects due to sparse point cloud data [15]. Several methods have overcome these limitations by fusing semantic information from the camera [16, 17, 18]. Though the RADAR-Camera fusion methods overcome the misclassification of objects, they cannot significantly recover the lost geometry due to a lack of depth perception capabilities [19]. Recent developments have shown significant improvements in object detection accuracy of RADAR-Camera-based systems through mid-level fusion but necessitate model training when employing a new camera with distinct intrinsic properties, hence complicating real-time application [20, 21].

On-board 3D object detection algorithms utilizing a single camera have demonstrated improvements in capturing the geometric information of faraway objects compared to LIDAR and RADAR-based methods [22, 23]. The deep learning and geometric constraint-based approach [22] comprises two phases: initially, it regresses the object's orientation and dimensions utilizing the deep learning model, followed by employing the 2D bounding boxes of the objects as constraints to determine their translation. The Single-Stage Monocular 3D Object Detection via Keypoint Estimation (SMOKE) method [23] utilizes keypoint estimation to regress the 3D parameters in one step directly. Although these techniques rely on on-board cameras, the deep learning model can be trained for roadside object detection by utilizing data from an infrastructure perspective. Nevertheless, similar to RADAR-Camera-based methods discussed previously, these methods necessitate model retraining when employing a new camera with distinct intrinsic properties, complicating real-time application.

MSight, an infrastructure-based perception system developed by the University of Michigan, utilizes fisheye cameras strategically located at the four corners of an intersection [24]. MSight employs a YOLO algorithm for detecting objects and converts image pixel coordinates into real-world coordinates via homography. The results demonstrate that the localization errors were evaluated to be approximately 1 meter; yet the method only provides 2D object information from the Bird's Eye View (BEV) perspective. 3D-Net and CARs On the Map (CAROM) are monocular camera-based traffic monitoring systems that provide three-dimensional information about objects [25, 26]. 3D-Net defines the midpoint at the bottom side of each 2D bounding box as the reference point for the objects. The pixel coordinates of the 2D image are transformed into real-world coordinates by homography. The heading angle of the objects is calculated using two real-world coordinates, x and y. When the object is stationary, the nearest road boundary is utilized as a reference to estimate the heading angle. The dimensions of the objects are determined based on the predefined dimensions of each object type. CAROM employs MaskRCNN and Sparse Optical Flow vectors as the foundational techniques for 3D object estimation [27, 28]. The optical flow vector converging at the vanishing point on the horizon line for each object is calculated using Random Sample Consensus (RANSAC) [29] and utilized for estimating the vehicle's heading angle. Vanishing points along the y-axis and z-axis are determined, and the three-dimensional bounding box is derived from the contour of its segmentation mask with the tangent line approach [30]. In this paper, the CAROM algorithm is chosen for roadside 3D object detection due to its practical implementation capability, which just requires camera calibration. Unlike other monocular object detection methods [22, 23], which often demand training with data from the specific camera used for implementation due to the varying intrinsic camera parameters involved, CAROM doesn't require explicit model training, enabling seamless real-world implementation. Moreover, the localization errors were determined to be approximately 0.8 meters within 50 meters of the camera and 1.7 meters within the range of 50 to 120 meters, which is considered to be within acceptable limits for practical implementation.

V2I technology involves fusion of data, features, or information from the RSU with the on-board unit. Based on the fusion stage, multi-agent cooperative perception includes three fusion schemes, namely, early, intermediate, and late fusion.

1. In early fusion, raw sensor data from multiple agents is fused before object detection [31, 32].
2. In intermediate fusion, intermediate features extracted from the feature extraction models are fused [33, 34, 35].
3. In late fusion, the object detection output from multiple agents is fused [36, 37].

In this paper, we propose a time delay compensation module to compensate for the processing and transmission delay from the roadside unit, followed by a simple Intersection over Union (IoU)-based object merger node to fuse object detections from multiple agents. Sections II-D and II-F provide detailed information on the time delay compensation module and the object merger node, respectively.

The literature review reveals that the majority of current V2I approaches utilize LIDAR, and even if it is a camera-based system, the model often requires re-training with data collected from the specific camera used. Moreover, most of the methods use data captured synchronously from multiple agents for fusion, thereby neglecting the processing and transmission delays experienced in practical implementations. Therefore, the aim of this paper is to develop a practically feasible V2I framework and demonstrate its effectiveness by testing and validating it in CARLA simulator [38].

The contributions of this paper are listed below:

1. Leverage the capabilities of an existing monocular object detection algorithm (CAROM), which could be implemented without the need for model retraining and could be used with any kind of monocular camera.
2. Develop a time delay compensation module to compensate for the processing and the communication latency from the roadside unit.
3. Implement a late-fusion method that fuses asynchronous object detection results from multiple agents.
4. Test and validate the proposed V2I framework by simulating a scenario similar to the one reported by Waymo [39].

The remainder of this paper unfolds as follows: Section II introduces the framework of the proposed V2I architecture; Section III provides a detailed explanation of the scenario used for evaluation; Section IV presents the validation and testing results; Section V includes some discussions on practical implementation of the framework; and finally, Section VI encapsulates the conclusion.

## II. PROPOSED V2I FRAMEWORK DESIGN

Fig. 1 illustrates the proposed V2I architecture, comprising two components: the RSU and the on-board unit of the CAV. The RSU component, which is the upper half of Fig. 1, consists of subcomponents such as a roadside camera, an edge computer, and a radio transmitter. The lower half represents the on-board unit. The on-board unit consists of both hardware and software components, which receive data from the RSU and process it to integrate it with the on-board perception system. The following subsections provide an in-depth summary of the subcomponents in the proposed V2I Framework, as shown in Fig. 1.

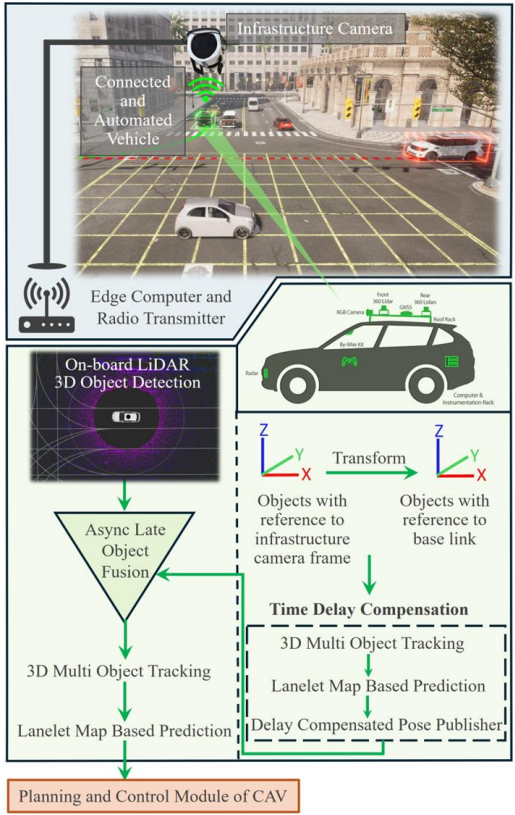

Fig. 1. V2I Framework Architecture

### A. Monocular Traffic Camera-based 3D Object Detection

For roadside perception, an RGB camera is strategically positioned to aim towards an intersection on the CARLA Town 10 map [28] as shown in Fig. 2a. We chose CAROM for object detection due to its practical implementation as mentioned in Section I. Camera calibration is a key component of CAROM, assuming the camera model is a pinhole camera with negligible lens distortion. To satisfy this assumption, the camera is calibrated for distortion parameters, and the images are rectified before performing the calibration to find the intrinsic and extrinsic camera parameters that are used to derive the homography matrix.

The calibration procedure constructs the camera projection matrix from the world frame (map frame) to the image frame, from which the homography matrix is derived. Fig. 2b and Fig. 2c illustrate the selected point correspondences between the local map frame and the image frame and their corresponding reference frames.

The following steps describe the calibration procedure:

1. Select at least six point correspondences between the image frame (in pixel coordinates) and the local map frame (in map pixel coordinates).

2. Transform local map pixels into the map frame. Equation 1 defines the transformation, wherein $P_{map}$ represents a point on the ground plane, $T_{L2M}$ represents the transformation between the local map pixel frame and the map frame in metric units, and $P_{local}$ denotes the point in the map pixel frame.

$$P_{map} = T_{L2M}.P_{local} \quad (1)$$

3. By utilizing the point correspondences between the map frame and the image plane, the camera projection matrix can be determined through the constrained least squares method. Equation 2 illustrates that A is a matrix of known values obtained from point correspondences, while P is a vector of unknown values that will be determined and subsequently rearranged into a 3x4 matrix that creates the projection matrix.

$$AP=0, \text{ where } ||P||^2 = 1 \quad (2)$$

4. Finally, the 3x3 homography matrix T is obtained by removing the vector along the z-axis from the projection matrix. This homography matrix is used to perform inverse perspective mapping (IPM), projecting the points from image pixel coordinates back to world coordinates from a bird's-eye perspective.

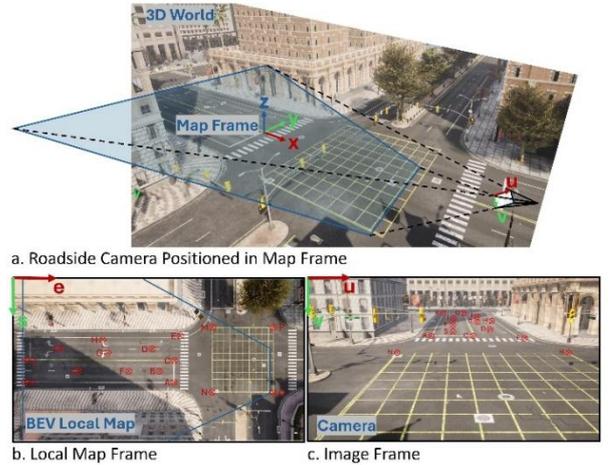

Fig. 2. Reference Frames in CAROM and the Point Correspondences Between the Local Map Frame and the Image Frame

CAROM uses MaskRCNN for object detection and instance segmentation on each image. The quality of the masks is crucial for accurate 3D bounding box detection, and hence the MaskRCNN is trained on a custom dataset with data collected from an infrastructure perspective. This ensures better performance compared to the pre-trained MaskRCNN. Sparse optical flow vectors are calculated for every object using the 2D Regions of Interest obtained from object detection. For each detected object, its vehicle type is recognized using a ResNet-18 [40] classifier trained on a custom dataset. The RANSAC algorithm is employed to find the optical flow vectors that converge at a vanishing point on the horizon, using which the heading angle of the vehicle is estimated. Using the center of the 2D bounding box, the homography matrix T and the heading angle, the other two vanishing points along the y-axis and z-axis are computed. Finally, with all three vanishing points, the 3D bounding box of a vehicle is computed from the contour of its segmentation mask using the tangent line method.

*B. RSU*

In real-world implementation, the RSU consists of an edge computer to process the camera image frames and a radio transmitter to transmit the processed information to the connected and automated vehicles (CAVs) nearby. Alternatively, as this paper focuses on simulation, modifications are made in the software to mimic the framework similar to a physical system. The object detection output from CAROM is transmitted to the CAV in the simulation only when the vehicle is within 80 m of the virtual RSU. This virtual RSU transmits both the object detection information and the processing time of that frame, which will be useful for the downstream pipeline during the late fusion process. Fig. 3 illustrates various reference frames and the transmission range of the virtual RSU within the simulator.

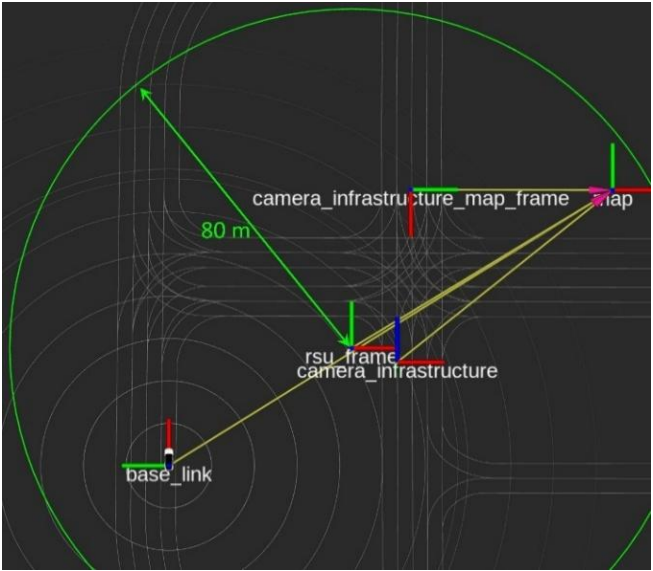

Fig. 3. Transmission Range of the Virtual RSU

*C. Object Frame Transformation*

Beginning this module, all the processes are handled by the on-board unit in the CAV. Once the on-board computer receives 3D object information from the RSU, the objects are transformed from the camera frame to the base link of the vehicle. To transform the objects from camera frame to base link, the transformation from camera frame to the base link of the CAV ($T_{Camera2Base}$) is needed, which is calculated using the transformation between camera frame and map frame ($T_{Camera2Map}$); and the base link of the CAV and the map frame ($T_{Map2Base}$). The $T_{Map2Base}$ is retrieved based on the camera image frame time as the reference time. Equation 3 defines the transformation between the camera frame and the base link.

$$T_{Camera2Base} = T_{Map2Base} \times T_{Camera2Map} \quad (3)$$

Using the $T_{Camera2Base}$ obtained from equation 3, the transformation between the object and the base link ($T_{Object2Base}$) is calculated based on equation 4. $T_{Object2Camera}$ is the transformation between the object and the camera frame and is a known value that is obtained from the RSU.

$$T_{Object2Base} = T_{Camera2Base} \times T_{Object2Camera} \quad (4)$$

Fig. 4 elucidates the various coordinate frames, and the transformation involved in finding the $T_{Camera2Base}$. The downstream pipeline receives the transformed objects for tracking and prediction, which are necessary to offset the CAROM processing delay and the message transmission delay.

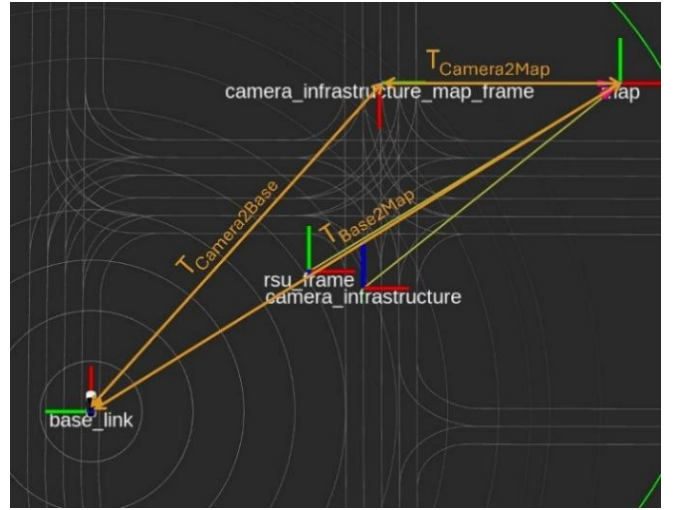

Fig. 4. Frame Transformations involved in Finding the Transformation between Camera Frame and Base Link

*D. Time Delay Compensation*

This module compensates for the delay from CAROM and data transmission from the RSU. It comprises three submodules, namely, tracking, prediction, and delay compensation.

Object tracking plays a vital role in predicting the future trajectory of objects. In this paper, muSSP: Efficient Min-cost Flow Algorithm is used for multi-object tracking (MOT) [41]. Existing approaches to solving min-cost flow problems in MOT are often suboptimal in terms of computational efficiency. Hence, the authors of the muSSP exploited the special structures and properties of graphs formulated in MOT problems to develop an efficient min-cost flow algorithm.

Following MOT, the future trajectories of objects are predicted using a lanelet map-based predictor. This paper employs a Frenét-frame-based method that uses a moving reference frame aligned with the road's centerline, which simplifies the trajectory planning by transforming it into a one-dimensional problem along the centerline and a perpendicular offset [42]. This module is configured to provide the predicted trajectories of the objects for the next 300 ms with a step size of 10 ms. Fig. 5 elucidates the output from this prediction module. This predicted trajectory is used to compensate for the delays caused by the CAROM and transmission.

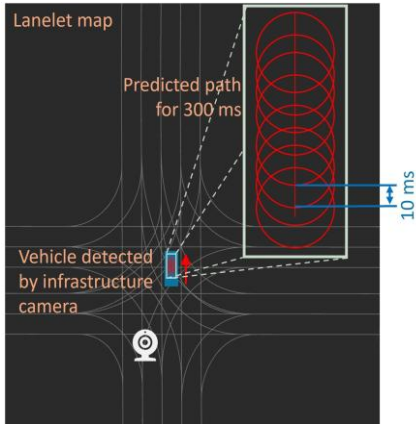

Fig. 5. Path Prediction Output from the Prediction Module

For each input camera frame with the image resolution of 896 x 504 pixels, the CAROM, on average, takes 80 to 120 ms to process and provide the output. This phased-out object detection information cannot be used for fusion with the object detection results from the on-board perception system. Moreover, CAROM has an output rate limit of 5 frames per second (fps), but the on-board sensors process data at 10 fps. Therefore, this module is designed to offset the time delay and also enhance the output rate of the results from the RSU. Firstly, the processing delay from CAROM is compensated by selecting a pose (first selected pose) from the predicted trajectory and publishing it to the downstream fusion module. Secondly, this module selects a second pose from the predicted trajectories to match the output rate of both sources. It publishes this second pose after the first selected pose, based on a pre-calculated time that also accounts for the processing delay from CAROM. In a physical system, the transmission delay is introduced at the beginning of the processing of the CAV's on-board system. However, to mimic the real-world transmission delay, a 10 ms delay is introduced in this module before publishing the poses to the downstream pipeline. The following pseudocode illustrates the algorithm clearly.

**Algorithm 1: Algorithm To Publish Predicted Poses**

1. **Set** input_frame_interval to 200 ms

    // Variable defined based on the time interval between camera frames.

2. **Set** prediction_point1 to the rounded value of CAROM_processing_time divided by 10

    // Set the first index value based on the processing time of CAROM. For example: If CAROM processing time is 110 ms, then the pose that needs to be published from the prediction is the value at index 11 (round (110/10) = 11).

3. **Delay** for 10 ms

    // To simulate for the transmission delay

4. **Increment** prediction_point1 by 1

    // The index is incremented by one which selects the next pose to the previously selected pose to compensate for the transmission delay that is manually created in the previous step.

5. **Publish** prediction_output at index prediction_point1

    // Publish the time delay compensated pose for the downstream pipeline.

6. **Set** start_time to current_time

7. **Loop** indefinitely

8.     Calculate wait_time as (input_frame_interval - rounded CAROM_processing_time) divided by 1000

    // Wait time calculated before publishing the second pose.

9.     **If** current_time - start_time is greater than wait_time then

10.         **Delay** for 10 ms

    // To simulate for the transmission delay

11.         **Set** predicted_point2 to 22

    # Set the second index value to 22 which corresponds to the pose at 220th ms and it includes the transmission delay as well.

12.         **Publish** prediction_output at index predicted_point2

13.     **End** if

14. **End** loop

15. **Jump** to Step 2 upon subscription of the next prediction result.

*E. On-board 3D Object Detection*

The test vehicle is equipped with a 64-channel LIDAR, positioned at the top center, with an output rate of 10 Hz. This module uses the LIDAR Center Point algorithm [43] for 3D object detection, which includes a two-stage detection process. In the first stage, a LIDAR-based backbone network such as VoxelNet [44] or PointPillars [45] is used to generate a representation of the input point cloud and convert to an overhead map view, using which the keypoint detector identifies object centers and the 3D object parameters are regressed. The second stage extracts point features at the 3D centers of each face of the estimated object's 3D bounding box and recovers the lost geometric information, which helps refine the regressed 3D object properties.

*F. Asynchronous Late Object Fusion*

This module is designed to fuse objects from the on-board perception system and the RSU. It is independent and works even without any messages received from the infrastructure unit. This module operates without requiring strict synchronization between agents. The only assumption is that agents share a common time reference, such as Unix time, and the time drift in the systems is corrected using Network Time Protocol periodically for better time synchronization between agents. When a message is received from both the on-board and infrastructure perception systems, the duplicate instances of the objects are filtered out and removed based on the overlap obtained from Intersection over Union (IoU) of the bounding boxes. If the same object is detected by both the on-board and infrastructure perception systems, the object instance from the infrastructure source is removed. If objects from different sources do not overlap, the system concatenates all the objects without any filtering. In the next phase, we plan on including filtering techniques such as the Kalman filter for data fusion, leveraging localization, and bounding box results from several sources instead of depending solely on a single source as it is in the current method.

### G. 3D MOT and Prediction

This module tracks and predicts the path of the fused objects from the RSU and on-board perception system. The MOT module uses muSSP: Efficient Min-cost Flow Algorithm, the same as the one used in the Time Delay Compensation module in Section II-D. Similarly, the prediction module uses lanelet map-based prediction, as in Section II-D. However, the parameters are tuned to predict the trajectories of the tracked objects for the next 10 seconds as expected by the planning and control module of the CAV [46].

## III. SCENARIO GENERATION IN CARLA SIMULATOR

The CARLA Scenario Runner [39] is a package that consists of several atomic behaviors and trigger conditions used for creating scenarios in the CARLA world. Scenario Runner supports scenario definition using various methods, namely the Python interface, OpenScenario, and OpenScenario 2.0. This paper utilizes the Python interface for scenario definition, as it integrates directly with Scenario Runner. For the scenario creation process, three fundamental behaviors known as atomic behaviors in CARLA, namely WaypointFollow, StopVehicle, and Wait, are employed. The WaypointFollow makes the vehicle follow a set of waypoints at a predefined speed, the StopVehicle behavior initializes braking and stops the vehicle, and the Wait behavior creates a delay of a few milliseconds between behavior changes for a seamless transition.

### A. Adversary Vehicle Running Red Light

According to the Waymo report [39], the other vehicle, which failed to stop at the red light and proceeded straight through the intersection, caused an actual collision to the side of the Waymo vehicle. The Waymo vehicle fails to detect the other vehicle running through the red light due to the occlusion caused by the adjacent lane vehicle. Fig. 6 illustrates the implementation of a similar scenario in the CARLA simulator through the use of CARLA Scenario Runner. The image depicts a projected collision if the AV goes straight, and the other vehicle runs through the red light (from left to right). However, in the simulated scenario, the AV is intentionally stopped at the stop line, though it has right of way, and the other vehicle is running through the red light. This helps demonstrate how the proposed V2I framework allows the AV to view the entire scene, despite the occlusion.

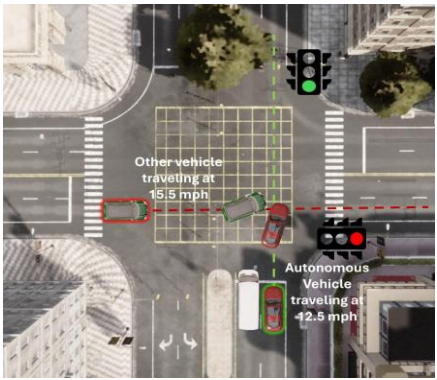

Fig. 6. CARLA Scenario: Vehicle Running Red Light

## IV. TESTING AND VALIDATION

The scenario described in Section III-A is utilized to evaluate the effectiveness of the proposed V2I framework. We test and validate the perception capabilities of the CAV both with and without V2I technology to demonstrate how the proposed system improves the CAV's perception through the integration of V2I technology.

### A. Evaluation of Time Delay Compensation Module

This subsection demonstrates the importance of the time delay compensation module and how it improves the localization accuracy of the infrastructure detections transmitted to the CAV. The vehicle running the red light mentioned in Section III-A is chosen for this evaluation. This evaluation involves two metrics: Mean IoU of the 2D bounding box of the object from the BEV perspective with the corresponding ground truth and the Root Mean Square Error (RMSE) of the localization results from BEV perspective.

The Mean IoU and RSME results of the object both with and without time delay compensation are provided in Table I. The results show an increase in mean IoU value and a decrease in RMSE, which demonstrates the improvement in localization accuracy due to time delay compensation. However, the RMSE results appear to be higher than expected due to the higher localization error of CAROM, as detailed in section II-A.

TABLE I. MEAN IoU AND RMSE RESULTS WITH AND WITHOUT TIME DELAY COMPENSATION

| Condition | Mean IoU | RMSE (meter) |
|---|---|---|
| Without time delay compensation | 0.21 | 3.32 |
| With time delay compensation | 0.34 | 2.52 |

The localization plots in Fig. 7 elucidate the improvement of object position in longitudinal direction with time delay compensation. But due to the lower output rate (5 Hz) of object detection results from the RSU, the offset in the lateral position has slightly increased compared to the lateral position without compensation. The lateral offset can be reduced with a higher output rate of at least 10 Hz from the RSU, which captures sudden changes in vehicle orientation.

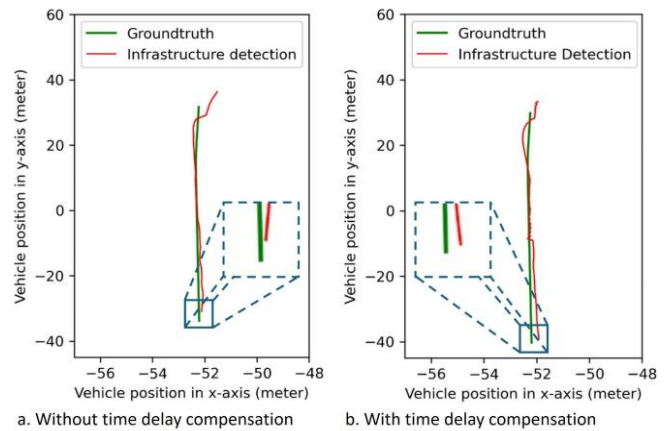

Fig. 7. Infrastructure Detection Result and the Corresponding Ground Truth in BEV perspective

## B. Evaluation of Object Fusion Module

Similar to the previous subsection, the asynchronous late object fusion module is evaluated using Mean IoU and RMSE. Fig. 8 shows the localization plot of the object after fusion, along with its corresponding ground truth location obtained from the CARLA Simulator [38]. The upper half of the detection is from the infrastructure unit, and the lower half of the detection is from the on-board sensor. Section II-F offers a comprehensive elucidation of the fusion process, which enhances the understanding of this fusion process.

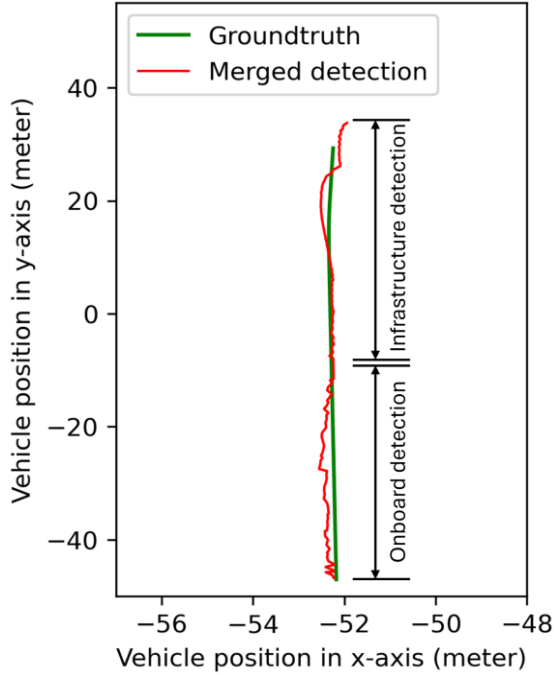

Fig. 8. Localization Plot of the Object After Late Fusion

The results presented in Table II demonstrate improvements in the localization accuracy following object fusion from both the RSU and the on-board unit. The Mean IoU is higher, and the RMSE is lower than the results from infrastructure-only detection. The improved accuracy is due to the fusion, where more than half of the detection is from the RSU and the remaining portion is from the on-board perception system, which has higher accuracy compared to the infrastructure unit. However, despite the RSU's lower accuracy compared to on-board perception, it significantly enhances the CAV's perception range.

TABLE II. MEAN IOU AND RMSE RESULT AFTER LATE FUSION

| Condition | Mean IoU | RMSE (meter) |
|---|---|---|
| Asynchronous late fusion with time delay compensation | 0.41 | 2.48 |

## C. Detection Range of the Other Vehicle by CAV

The results from Section IV-B are further supported by calculating the CAV's detection distance of the other vehicle using both vehicle-only perception and V2I fusion, as presented in Table III. Fig. 9 illustrates that when using vehicle-only perception, the CAV only detects the other vehicle after it has entered the intersection, potentially leading to a crash. However, when using V2I fusion, the CAV can detect the other vehicle even before it enters the intersection, providing ample time and distance for the CAV to react.

TABLE III. DETECTION RANGE OF OTHER VEHICLE BY CAV

| Distance between CAV and other vehicle on first detection | Euclidean distance (meter) |
|---|---|
| With vehicle-only | 21.6 |
| With V2I late fusion | 76.0 |

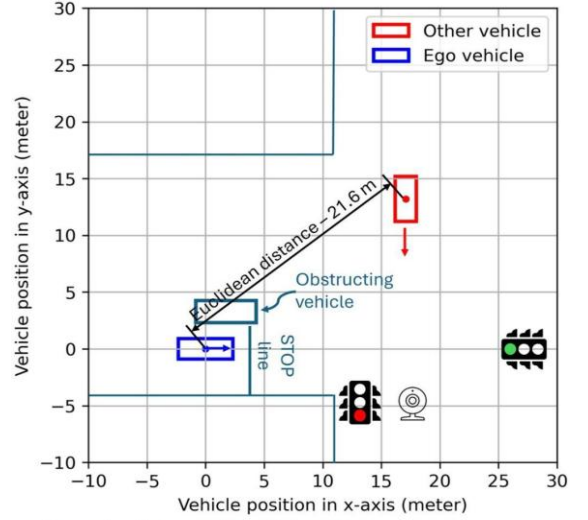

a. Detection distance with vehicle-only

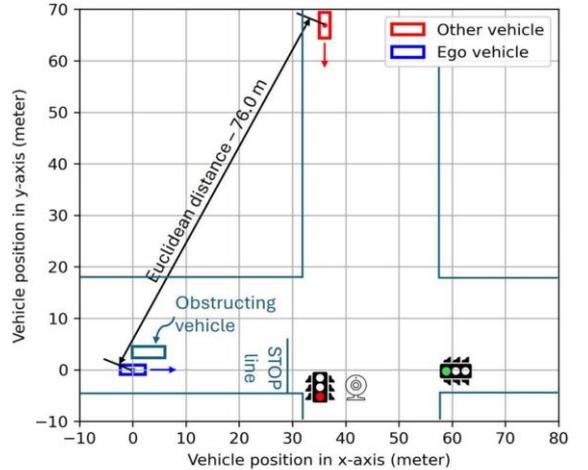

b. Detection distance with V2I Fusion

Fig. 9. Detection distance of the other vehicle by CAV

## V. DISCUSSION ON PRACTICAL IMPLEMENTATION

The purpose of this section is to discuss some criteria that need to be considered in real-world applications of the proposed V2I framework.

### A. Estimation of Time Delay of Messages from RSU

This subsection provides the method used in this paper to find the time delay of the messages from the RSU and the approach/modification required to estimate the time delay in order to achieve real-world implementation of the framework. As mentioned in Section II-B, the virtual RSU transmits both the object information and the image processing time, which is utilized by the time delay compensation module to offset the delay. In addition to the processing delay, a constant transmission delay of 10 ms is assumed and compensated. However, in practical implementation, we cannot rely on the processing time transmitted from the RSU, and moreover, the transmission delay is not consistent and varies based on CAV's distance from the RSU. Therefore, in a practical scenario, the time difference between the message from the on-board perception system and the RSU must be calculated in the on-board computing unit. The calculated time difference includes both the processing time and transmission delay from RSU, using which the future pose can be estimated. With improved output rate from RSU, the new approach of calculating the time delay will be utilized in the next phase of the project.

### B. Detection Range and Processing Speed of Roadside Perception System

This subsection provides some insight into how the resolution of the camera can affect the detection range and processing speed of CAROM. As previously mentioned in Section II-A, CAROM employs MaskRCNN and Sparse Optical Flow vectors to estimate the 3D bounding box parameters. The accuracy of MaskRCNN and Sparse Optical Flow vectors heavily relies on the resolution of the camera. Sharper images enhance the performance of these algorithms, enabling them to segment and estimate the optical flow vectors of distant objects more accurately, thereby expanding the detection range. However, the higher detection range comes with a trade-off of higher computational cost and lower processing speed. Additionally, the image processing time varies depending on the number of vehicles present in the frame. Therefore, the camera's output rate should match the longest time the image processing algorithm needs to handle an image, regardless of the number of vehicles. For instance, if it takes 200 milliseconds to process one image, the camera's output rate should be configured such that it captures frames at intervals of at least 200 milliseconds (i.e., a maximum frame rate of 5 fps). This guarantees the processing of each frame before the arrival of the next. Hence, real-world applications require careful selection and configuration of hardware resources to achieve the required detection range and accuracy with a significant output rate. In this implementation, we were able to achieve a maximum detection range of 80 meters and an output rate of 5 fps with a camera resolution of 896 x 504 pixels while running on a Nvidia RTX 4070 laptop GPU.

### C. Operational Design Domain

Operational Design Domain (ODD) plays a major factor in real-world implementation. The current iteration of the framework is tested under normal daylight conditions, based on which CAROM is designed. However, practical implementation requires a wider operational domain, including varying weather conditions such as cloudy and rainy. Upcoming improvements will focus on utilizing an IR vision-based camera system, facilitating broader ODD. Another important consideration is the detection and tracking of high-speed vehicles and Vulnerable Road Users (VRUs). The current implementation has limitations on real-time tracking of high-speed vehicles and VRUs due to lower processing speed and semantic segmentation accuracy, respectively. To support high-speed applications and detection of VRUs, upcoming work will focus on enhancing the processing speed and segmentation accuracy through state-of-the-art semantic segmentation algorithms, making the framework robust for practical implementation.

## VI. CONCLUSION

This paper introduces a V2I asynchronous late object fusion framework leveraging the capabilities of roadside cameras, designed explicitly to increase the perception capability of CAVs through enhanced scene representation at a lower implementation cost. The proposed time delay compensation module improves the synchronization of the results from the roadside unit with the on-board perception system of CAV, revealing it as a viable solution for the deployment of cooperative driving systems in real-world conditions. The simulation results presented herein highlight the effectiveness of the framework and its potential for enhancing the safety and reliability of automated driving solutions. Upcoming improvements will focus on enhancing the Operational Design Domain of the framework, focusing on varying weather conditions, detection of high-speed vehicles, and VRUs.

DEFINITIONS/ABBREVIATIONS

**BEV** - Bird's Eye View

**CAROM** - CARs On the Map

**CAV** - Connected and Automated Vehicle

**IoU** - Intersection over Union

**MOT** - Multi-Object Tracking

**RANSAC** - RANdom SAmple Consensus

**RMSE** - Root Mean Square Error

**RSU** - Roadside Unit

**V2I** - Vehicle-to-Infrastructure

**VRU** - Vulnerable Road User